\documentclass[10pt,twocolumn,letterpaper]{article}
\usepackage{cvpr}              
\usepackage[accsupp]{axessibility}  

\definecolor{cvprblue}{rgb}{0.21,0.49,0.74}
\usepackage[bookmarks=false,pagebackref,breaklinks,colorlinks,allcolors=cvprblue]{hyperref}

\usepackage{multirow}
\usepackage{amsthm}
\usepackage{float}
\usepackage[T1]{fontenc}

\def\paperID{40} 
\def\confName{CVPR}
\def\confYear{2025}

\definecolor{somegray}{rgb}{0.5, 0.5, 0.5}
\newcommand{\darkgrayed}[1]{\textcolor{somegray}{#1}}
\makeatletter
\newcommand*\titleheader[1]{\gdef\@titleheader{#1}}
\AtBeginDocument{%
  \let\st@red@title\@title
  \def\@title{%
    \vskip-3em
    \bgroup\normalfont\large\centering\@titleheader\par\egroup
    \vskip1.5em\st@red@title}
}
\makeatother

\titleheader{\darkgrayed{This paper has been accepted for publication at the \\
IEEE Conference on Computer Vision and Pattern Recognition Workshops (CVPRW), Nashville, 2025.
\copyright IEEE}}

\title{Perturbed State Space Feature Encoders for Optical Flow with Event Cameras}

\author{Gokul Raju Govinda Raju$^{*}$ \quad Nikola Zubi\'{c}$^{*}$ \quad Marco Cannici \quad Davide Scaramuzza\\Robotics and Perception Group, University of Zurich\\{\tt\small ggovinda@student.ethz.ch, \{zubic, cannici, sdavide\}@ifi.uzh.ch}\\$^{*}$Equal contribution}

\begin{document}
\maketitle
\begin{abstract}
With their motion-responsive nature, event-based cameras offer significant advantages over traditional cameras for optical flow estimation. While deep learning has improved upon traditional methods, current neural networks adopted for event-based optical flow still face temporal and spatial reasoning limitations. We propose Perturbed State Space Feature Encoders (P-SSE) for multi-frame optical flow with event cameras to address these challenges. P-SSE adaptively processes spatiotemporal features with a large receptive field akin to Transformer-based methods, while maintaining the linear computational complexity characteristic of SSMs. However, the key innovation that enables the state-of-the-art performance of our model lies in our perturbation technique applied to the state dynamics matrix governing the SSM system. This approach significantly improves the stability and performance of our model. We integrate P-SSE into a framework that leverages bi-directional flows and recurrent connections, expanding the temporal context of flow prediction. Evaluations on DSEC-Flow and MVSEC datasets showcase P-SSE's superiority, with 8.48\% and 11.86\% improvements in EPE performance, respectively.
\end{abstract}
\section{Introduction}
\label{sec:intro}
Optical flow estimation, a cornerstone of computer vision, plays a crucial role in diverse applications such as object tracking \cite{Serych_2023_WACV, kiran_kale}, robotic navigation \cite{qin_tong, Min_2020_CVPR}, video interpolation \cite{Tulyakov_2021_CVPR, Tulyakov_2022_CVPR, Kim_2023_CVPR}, and action recognition \cite{Piergiovanni_2019_CVPR, Sun_2018_CVPR}. Recent deep learning approaches \cite{Teed2020RAFTRA, jahediMultiScaleRAFTCombining2022, Shi_2023_ICCV} have significantly outperformed traditional energy minimization techniques \cite{horn_1981}, advancing the field substantially.

However, optical flow estimation with traditional frame-based cameras faces challenges because images are captured at fixed time intervals, leading to temporal gaps, which could affect motion information. To overcome these limitations, researchers have turned to alternative imaging devices like event-based cameras \cite{DBLP:journals/corr/abs-1904-08405}, which provide a continuous, asynchronous stream of data describing changes in relative pixel brightness with high temporal precision.

Despite having access to rich and continuous motion information, existing learning-based optical flow estimation methods that leverage event-based data, \cite{Gehrig3dv2021,Gehrig2024pami, Zhu-RSS-18, Luo_2023_ICCV} are still limited to pair-wise network designs, where only the local neighborhood of events surrounding two consecutive time instants is used to predict the corresponding optical flow. As a result, valuable temporal cues from preceding and successive instants are ignored.
On the other hand, recent state-of-the-art methods in frame-based optical flow, such as VideoFlow \cite{Shi_2023_ICCV}, have demonstrated that a multi-frame approach that optimizes bidirectional flows across multiple frames leads to superior performance, exploiting richer latent information within extended frame sequences. 

While VideoFlow \cite{Shi_2023_ICCV} demonstrates improved temporal reasoning, it faces two significant limitations. First, the inherent temporal resolution constraints of traditional cameras hinder their ability to capture fine-grained motion details. Second, its reliance on the Twins Transformer \cite{chu2021Twins} architecture for encoding introduces spatial limitations. The lack of a global effective receptive field in this encoder restricts the network's capacity to capture extensive spatial dependencies. Consequently, these constraints potentially impair VideoFlow's ability to effectively handle scenarios involving occluded objects or large-scale motions, highlighting areas where further advancements could yield substantial improvements in optical flow estimation.

Drawing inspiration from recent advancements in frame-based optical flow techniques, we present a novel multi-event optical flow estimation approach. Our model departs significantly from conventional event-based methods through two key innovations. First, we implement bi-directional optical flow prediction and optimization across up to five consecutive time instants. This allows us to recursively propagate motion information between adjacent frames, substantially expanding the network's temporal receptive field. To address the spatial limitations of existing methods, we introduce the Perturbed State Space Feature Encoders (P-SSE) to optical flow estimation—a first in this domain. This novel encoder is designed to capture more comprehensive spatial dependencies, improving the model's ability to handle complex scenarios. These innovations combine to leverage extended temporal context and enriched spatial information, significantly improving optical flow estimation accuracy and robustness.

The key innovation of P-SSE lies in the novel perturbation technique applied to the state dynamics matrix. While P-SSE builds upon the strengths of Transformers and Convolutional Neural Networks (CNNs), incorporating dynamic weights and global receptive fields characteristic of models like DeiT \cite{pmlr-v139-touvron21a}, its distinguishing feature is the perturbation method that crucially improves its performance and stability.
The perturbation technique is the key idea of P-SSE's capabilities, making the state dynamics matrix of the State Space Model (SSM) more robust to noise and significantly stabilizing the entire system. By applying a carefully designed perturbation followed by diagonalization, we address the inherent instability issues often associated with SSMs, particularly when processing complex, asynchronous event data.
P-SSE utilizes the scanning procedure of the VMamba model \cite{liu2024vmamba}. Still, our perturbation approach regularizes the state matrix and improves stability, allowing the model to capture the spatial context in event data more effectively. This improved spatial reasoning, combined with temporal insights from our VideoFlow-inspired \cite{Shi_2023_ICCV} multi-event strategy, results in a comprehensive spatiotemporal approach to optical flow estimation.
This innovative combination of perturbed spatial processing and multi-event temporal processing allows our model to capture complex motion patterns with strong accuracy and robustness, setting new standards in event-based optical flow estimation.

We conducted a comprehensive evaluation of our model on the DSEC-Flow \cite{Gehrig21ral} and MVSEC \cite{zhu_ral_2018} datasets, revealing several contributions:
\begin{itemize}
\item Our experiments demonstrate the successful adaptation of multi-frame optical flow strategies to event-based vision, resulting in substantial performance improvements.
\item The global receptive field and perturbation technique provided by our proposed P-SSE, combined with the model's capacity to leverage rich temporal motion information, significantly improves performance in challenging scenarios characterized by motion blur and occlusions.
\item On the DSEC-Flow benchmark \cite{Gehrig21ral}, our method achieves a top EPE score of 0.680, representing an 8.48\% improvement over the previous best TMA model \cite{Liu_2023_ICCV}, and a 9.33\% improvement compared to the BFlow \cite{Gehrig2024pami} model. On the MVSEC dataset \cite{zhu_ral_2018}, we observe an 11.86\% increase in EPE performance.
\end{itemize}
\section{Related Work}\label{sec:related_work}
\paragraph{Event-based optical flow}
Event-based optical flow estimation has evolved through several distinct phases, each adapting to the unique characteristics of event data. Initial approaches focused on modifying classical computer vision techniques, including adaptations of the Lucas-Kanade algorithm \cite{BenosmanICBS12,gehrig2020eklt,orchard2013spiking}, local plane fitting strategies \cite{benosman2013event,mueggler2015lifetime}, and methods recovering pixel trajectories via contrast maximization on the image plane \cite{gallego2018unifying}.
The success of learning-based approaches in frame-based optical flow catalyzed a shift towards adapting these architectures for event processing. This transition necessitated the development of structured representations of raw event data \cite{Gehrig3dv2021, Liu_2023_ICCV, Zubic_2023_ICCV}, capable of encapsulating rich motion details. Concurrently, unsupervised methods emerged, estimating sparse optical flow by minimizing photometric loss over warped images \cite{Zhu-RSS-18} or event-frames \cite{ye2018unsupervised}, often incorporating additional smoothness constraints \cite{yu2016back} or adversarial losses \cite{lai2017semi}.
Recent advancements have prioritized dense predictions, aiming to provide reliable estimates even in event-free regions. E-RAFT \cite{Gehrig3dv2021} pioneered this approach by adapting the RAFT architecture \cite{Teed2020RAFTRA}, utilizing correlation volumes and iterative refinement for forward optical flow computation in event data. Building on this foundation, TMA \cite{Liu_2023_ICCV} introduced a pattern aggregation mechanism to enhance motion features and improve convergence. Concurrently, ADMFlow \cite{Luo_2023_ICCV} addressed the challenge of varying event densities through an adaptive event representation, significantly improving robustness.
Despite these advancements, current event-based optical flow methods face limitations in both temporal and spatial reasoning. The prevalent reliance on pairwise predictions restricts the temporal context. In contrast, using encoders with limited spatial receptive fields that cannot stably encode rich contexts constrains the ability to capture and encode global motion patterns.
Our research diverges from previous approaches by adopting a VideoFlow-inspired \cite{Shi_2023_ICCV} architecture to exploit the rich temporal information inherent in multi-event sequences, enabling precise bidirectional optical flow predictions. This capability is further supported by integrating P-SSE, which efficiently captures long-range spatial dependencies via global context. Most recent event-based vision works include IDNet \cite{idnet}, which employs a lightweight event representation and recurrent refinement for robust flow predictions. Also, ECDDP \cite{Yang2024EventCD} applies a cross-scale dynamic propagation approach to capture finer spatiotemporal details. Although these methods were released after our initial submission, we report better results and include a discussion and comparison in our results section to ensure a comprehensive evaluation of recent state-of-the-art.

\paragraph{State Space Models}
State Space Models (SSMs) represent a novel class of sequence models in deep learning, bridging RNNs, CNNs, and classical state space models \cite{gu2021combining, gu2022efficiently, smith2023simplified, zubic2025limitsdeeplearningsequence, zubic2024ggssmsgraphgeneratingstatespace, soydan2024s7selectivesimplifiedstate}. Inspired by continuous state space models from control systems and utilizing HiPPO initialization \cite{gu2020hippo}, LSSL \cite{gu2021combining} showed promise in addressing long-range dependencies. However, LSSL's practical application was limited by its computational and memory intensity. S4 \cite{gu2022efficiently} addressed this by introducing parameter normalization into a diagonal format, enabling the development of various deep SSM architectures. These include complex-diagonal models \cite{dssgupta2022, gu2022s4d}, multi-input multi-output support \cite{smith2023simplified}, diagonal plus low-rank decompositions \cite{liquids4hasani2022liquid}, and selective mechanisms like Mamba \cite{mamba}. This evolution of SSMs has significantly expanded their applicability and efficiency in handling complex sequential data, making them particularly relevant for our work in event-based optical flow estimation.

In the domain of visual applications, S4ND \cite{s4ndnguyen2022s4nd} pioneered using SSMs. However, its direct extension of the S4 framework fell short in dynamically capturing image-specific details, largely due to its time-invariant nature. This limitation also manifests in the initial implementation of SSMs for event-based vision \cite{Zubic_2024_CVPR}, which employed a similar recurrent, time-invariant SSM approach for object detection. While drawing inspiration from Vim \cite{vim}, and VMamba \cite{liu2024vmamba}, our work goes beyond their capabilities by introducing a more stable system state through the novel perturbation technique. This advancement enables P-SSE to encode the spatial context of motion patterns more effectively and over longer sequences.

\section{Methodology}
Our approach is influenced by the latest advances in multi-frame optical flow estimation for image datasets \cite{Ferede_2023, Ren_2019}, particularly the VideoFlow model \cite{Shi_2023_ICCV}, which we adapt for event-based vision. Explained in Section \ref{sec:model_arch}, we exploit VideoFlow's TRi-frame Optical Flow (TROF) module, which operates on three frames, and its extension, the MOtion Propagation (MOP) module, which interconnects multiple TROF over five frames. While traditional models have focused on analyzing pairs of consecutive event representations to compute optical flows, leveraging the high temporal resolution of event data, they often overlook the valuable temporal insights available from adjacent events. Inspired by VideoFlow, our architecture employs a longer temporal context spanning up to five consecutive event representations, exploiting the rich motion information inherent in event data, thereby improving the precision of our optical flow estimations. Moreover, as explained in Section \ref{sec:tv_sse}, the spatial encoding in our architecture benefits from a global effective receptive field at linear computational expense thanks to the P-SSE. With its novel perturbation technique, the P-SSE plays a crucial role in our methodology by significantly improving the stability and performance of the state dynamics matrix, enabling our model to capture complex spatial-temporal dependencies in event data with consistent improvements in accuracy and efficiency.

\subsection{Problem Formulation for Event-based Vision}\label{sec:prob}
Optical flow estimation aims at computing a displacement field $\mathbf{f}_{i\rightarrow i+1}\colon \mathbb{I}^{H\times W \times 2} \rightarrow \mathbb{R}^{H\times W \times 2}$ mapping each pixel's coordinates from an initial timestamp $t_i$ to their new positions at a subsequent timestamp $t_{i+1}$, which we refer to as the forward optical flow. Conversely, the backward optical flow, $\mathbf{f}_{i+1{\rightarrow}i}$, maps pixel positions from $t_{i+1}$ back to $t_i$. 

\subsection{Model Architecture}\label{sec:model_arch}

Denote the stream of events captured by an event-based camera between two instants $t_h$ and $t_k$ as $\mathcal{S}(t_h,t_k) = \langle(\mathbf{e}_i|t_i\in[t_h,t_k] \rangle$. Each event is represented by a tuple $\mathbf{e}_i = \left(x_i, y_i, t_i, p_i\right)$, where $\left(x_i, y_i\right)$ specifies the pixel's location, the time of the event $t_i$, and the polarity $p_i$, indicating whether the brightness increased or decreased. 

Given a sequence of successive timestamps $\{t_{i-2}, t_{i-1}, t_{i}, t_{i+1}, t_{i+2}\}$ among which to estimate the optical flow, we first aggregate the events triggered in between these timestamps into a sequence of image-like representations $\mathcal{E}_{i} \in \mathbb{R}^{H \times W \times C}$, each one aggregating the events $\mathcal{S}(t_{i-1},t_i)$ into $C$-dimensional pixel-wise features. In this work, we employ ERGO-12 \cite{Zubic_2023_ICCV} and time surfaces \cite{Sironi18cvpr} due to their favorable balance between complexity and performance \cite{Zubic_2023_ICCV}. Then, following \cite{Shi_2023_ICCV}, we perform hierarchical recurrent processing, where we first process these representations in overlapping triplets $\{\mathcal{E}_{i-2},\mathcal{E}_{i-1},\mathcal{E}_{i}\}$, $\{\mathcal{E}_{i-1},\mathcal{E}_{i},\mathcal{E}_{i+1}\}$ and $\{\mathcal{E}_{i},\mathcal{E}_{i+1},\mathcal{E}_{i+2}\}$ using the Event-Triplet Optical Flow (E-TROF) module, and then interconnect them through the Event-Motion Propagation (E-MOP) module. 

\begin{figure}[t]
    \centering
    \includegraphics[width=\columnwidth]{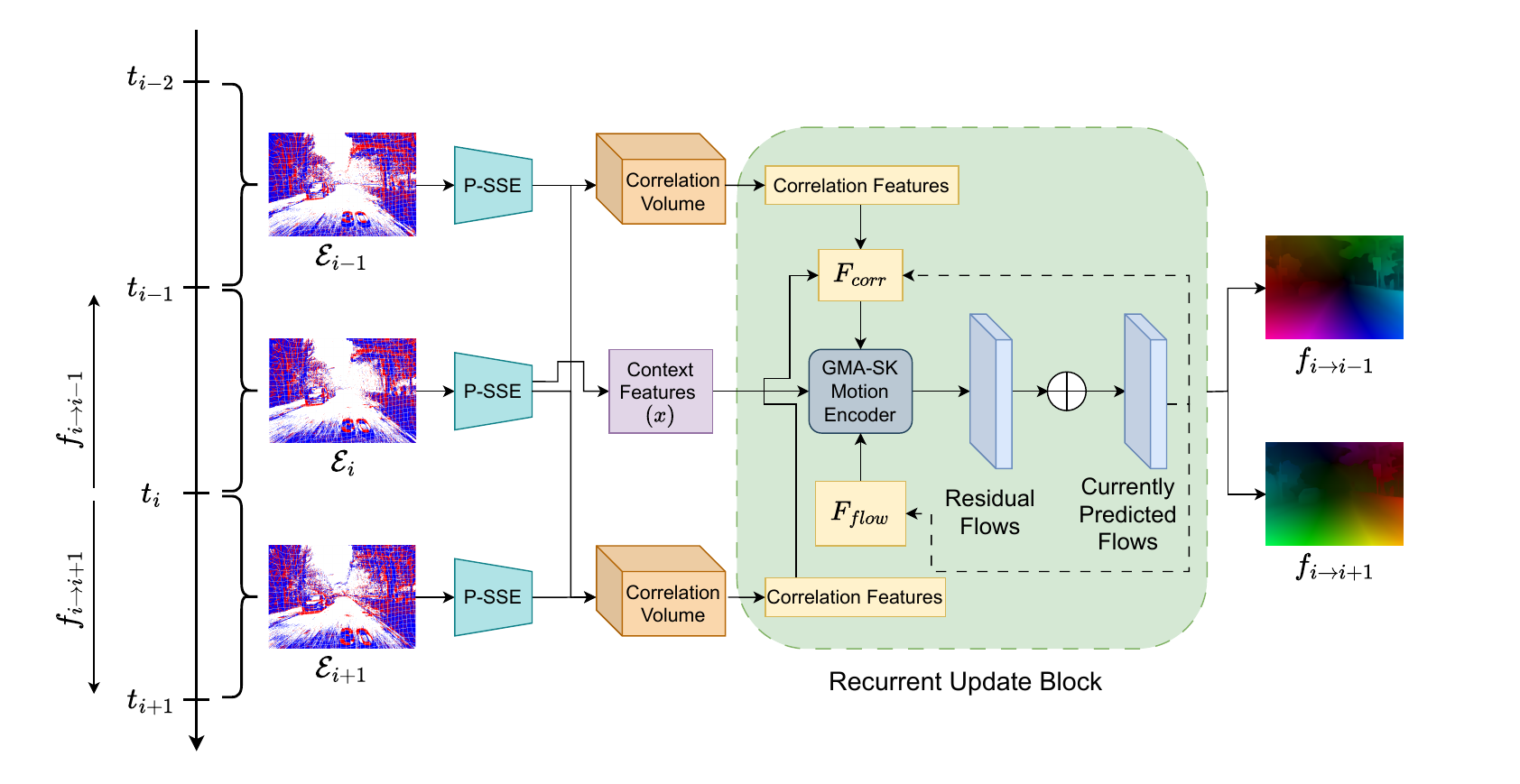}
    \caption{Illustrative diagram of our E-TROF model. Utilizing three successive event representations, E-TROF employs bidirectional correlation, context, and flow features for recurrently estimating bidirectional optical flows.}
    \label{img:etrof}
\end{figure}

\subsubsection{E-TROF (Event-Triplet Optical Flow)} \label{subsubsec:etrof} As depicted in Figure \ref{img:etrof}, the E-TROF module takes a triplet $\mathcal{E}_{i-1}$, $\mathcal{E}_i$, and $\mathcal{E}_{i+1}$ and predicts the bidirectional optical flows $\mathbf{f}_{i{\rightarrow}i+1}$ and $\mathbf{f}_{i{\rightarrow}{i-1}}$ centered around $t_i$. The process starts with a feature encoding phase, where event representations are transformed into feature embeddings through the proposed P-SSE model, explained in detail in Section \ref{sec:tv_sse}, which extracts features at 1/8th the original resolution of the event data. Following this, we construct forward and backward correlation volumes, $\mathbf{Corr}_{i,i+1}$ and $\mathbf{Corr}_{i,i-1}$, encoding the similarities between pixels across frames, to serve as the basis for deducing both forward and backward optical flows. The forward correlation volume is derived from $\mathcal{E}_i$ and $\mathcal{E}_{i+1}$, while the backward volume from $\mathcal{E}_i$ and $\mathcal{E}_{i-1}$. Context features $\mathbf{x}$, encoded from $\mathcal{E}_i$ using the same exact encoder, complement these correlation volumes.

Subsequently, we utilize the forward and backward correlation features to derive the combined bidirectional correlation features. This is achieved through an interactive process, similar to RAFT \cite{Teed2020RAFTRA}, where forward and backward flows are initialized as zero-displacements, $\mathbf{f}^0_{i{\rightarrow}i+1} = \mathbf{0}$ and $\mathbf{f}^0_{i{\rightarrow}i-1} = \mathbf{0}$, and then progressively refined over $K$ iterations:
\begin{gather}\label{eq:fusion}
    \mathbf{c}^k_{i{\rightarrow}i-1} = \mathbf{Corr}_{i,i-1}(\mathbf{f}^k_{i{\rightarrow}i-1}),\\
    \mathbf{c}^k_{i{\rightarrow}i+1} = \mathbf{Corr}_{i,i+1}(\mathbf{f}^k_{i{\rightarrow}i+1}),\\
    \mathbf{F}^k_{corr} = CorrelationFusion\left(\mathbf{c}^k_{i{\rightarrow}i-1}, \mathbf{c}^k_{i{\rightarrow}i+1}\right),\\
    \mathbf{F}^k_{flow} = FlowFusion\left(\mathbf{f}^k_{i{\rightarrow}i-1}, \mathbf{f}^k_{i{\rightarrow}i+1}\right).
\end{gather}

where correlation values $\mathbf{c}^k$ are obtained by sampling the corresponding correlation volumes based on the predicted flows, while flow features $\mathbf{F}^k_{corr}$ and $\mathbf{F}^k_{flow}$ through learnable encoders.
Following the fusion step, we compute the motion features using the Super Kernel Motion Encoder \cite{Shi_2023_ICCV}, which takes as input the fused bidirectional correlation $\mathbf{F}^k_{corr}$, the flow features $\mathbf{F}^k_{flow}$, and an additional feature map $\mathbf{F}^k_{map}$, described later in Section \ref{subsubsec:emop}, which carries long-range motion features from adjacent triplets:

\begin{equation}\label{eq:motionenc}
\begin{split}
    \mathbf{F}^k_{motion}, \mathbf{M}^{k+1}_i &= SK\text{-}MotionEncoder\left(\mathbf{F}^k_{corr}, \right.\\
    &\left. \mathbf{F}^k_{flow}, \mathbf{F}^k_{mop}\right),
\end{split}
\end{equation}
where $\mathbf{M}^{k+1}_i$ is a feature encoding the motion state at the center instant $t_i$.

The flow refinement process is analogous to RAFT \cite{Teed2020RAFTRA}, but carried over for both the forward and backward flow simultaneously, leveraging a hidden state feature $\mathbf {h}^k$, initialized with the context feature $\mathbf{x}$:
\begin{gather}\label{eq:updateblock}
\mathbf{h}^{k+1} = SK\text{-}UpdateBlock\left(\mathbf{F}_{motion}, x, \mathbf{h}^{k}\right),\\
\Delta{\mathbf{f}^{k}_{i{\rightarrow}i-1,i{\rightarrow}i+1}} = FlowDecoder\left(\mathbf{h}^{k+1}\right),\\
\label{eq:updateflow}
\mathbf{f}^{k+1}_{i{\rightarrow}i-1,i{\rightarrow}i+1} = \mathbf{f}^{k}_{i{\rightarrow}i-1,i{\rightarrow}i+1} + \Delta{\mathbf{f}^{k}_{i{\rightarrow}i-1,i{\rightarrow}i+1}}
\end{gather}

The flow estimates are similarly scaled because the feature embeddings were initially computed at a resolution reduced to 1/8th of the original event data. The final step involves upscaling the bidirectional flow predictions to match the original resolution of the event data.

\subsubsection{E-MOP (Event-Motion Propagation)} \label{subsubsec:emop} 
\begin{figure*}[t]
    \centering
    \includegraphics[width=0.7\textwidth]{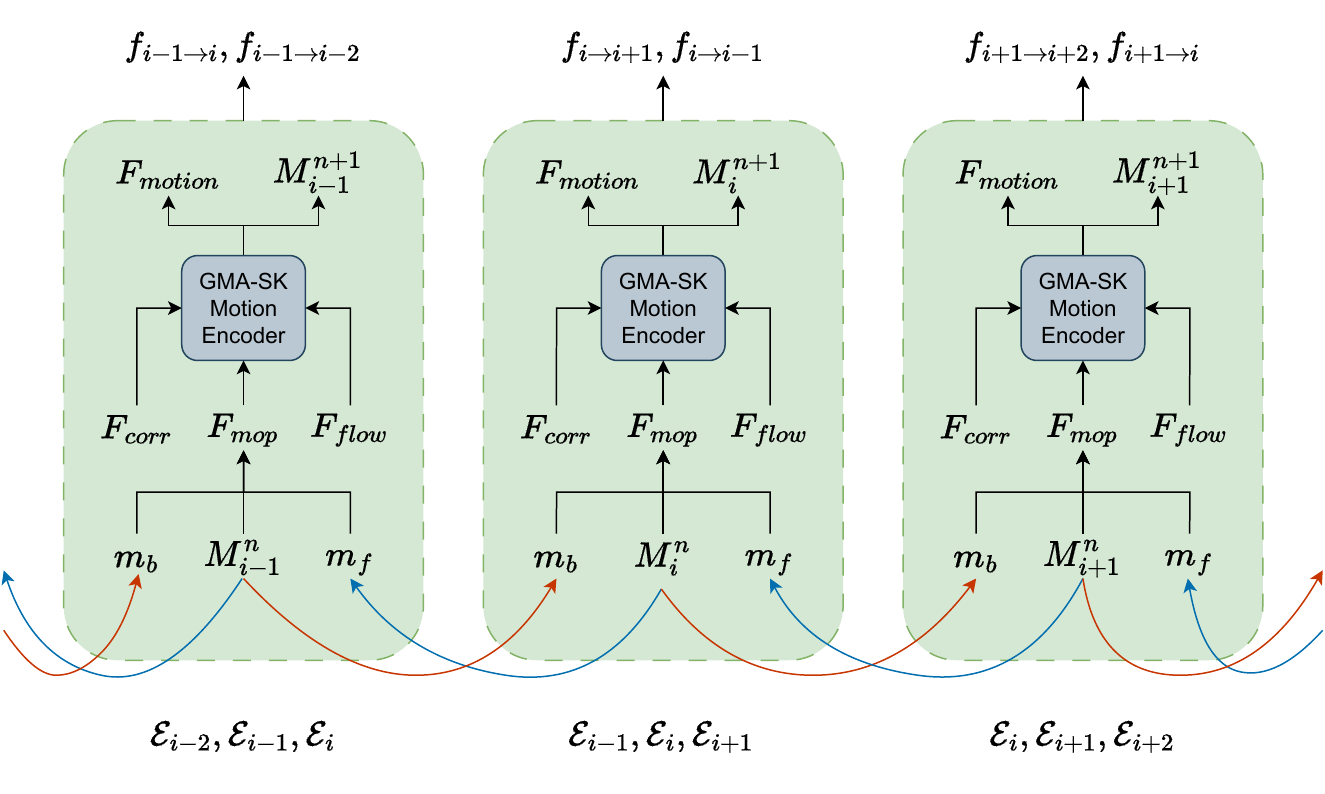}
    \caption{Diagram of the E-MOP model, which integrates 3 E-TROFs for 5 consecutive event representations to predict and refine bidirectional optical flows by sharing dynamic temporal motion information among adjacent TROFs.}
    \label{img:emop}
\end{figure*}

The E-MOP propagates motion information across triplets of event frames, enabling the exchange of motion state information between them to collaboratively refine the bidirectional optical flows estimated for each E-TROF segment, as illustrated in Figure \ref{img:emop}. To do so, a motion state feature $\mathbf{M}^{k}_i$, initially set to a random value, is computed from each triplet of event frames and then progressively refined. 
A bidirectional temporal motion information is extracted by warping the motion state features from adjacent TROFs, $\mathbf{M}_{i-1}$ and $\mathbf{M}_{i+1}$, according to the predicted optical flows and the current TROF's motion state feature.
\begin{gather}\label{eq:mop}
\mathbf{m}^k_{f} = Warp\left(\mathbf{M}^k_{i+1}, \mathbf{f}^k_{i\rightarrow{i+1}}\right),\\
\mathbf{m}^k_{b} = Warp\left(\mathbf{M}^k_{i-1}, \mathbf{f}^k_{i\rightarrow{i-1}}\right),\\
\mathbf{F}^k_{mop} = Concatenate\left(\mathbf{M}_{i}, \mathbf{m}_{f}, \mathbf{m}_{b}\right).
\end{gather}
Following this, the estimated motion features are first used to extract motion features as in Equation \eqref{eq:motionenc} and then used to refine the bidirectional flows iteratively through Equation \eqref{eq:updateflow}.
The continuous update of the motion state feature allows each event triplet to benefit from the temporal insights of adjacent triplets, enhancing the temporal depth and exploiting evolving temporal cues through iterative refinement.

\subsection{Perturbed State Space Feature Encoder (P-SSE)}\label{sec:tv_sse}
Our P-SSE encoder leverages state space models to achieve an effective global receptive field while maintaining linear complexity. At the heart of our encoder is a linear system that maps a 1D input signal of length $L$, $x(t) \in \mathbb{R}^{L}$, to an output response $y(t) \in \mathbb{R}^{L}$ through a hidden state $h(t) \in \mathbb{R}^N$:
\begin{equation}\label{eq:ssm}
\begin{split}
  h'(t) = \mathbf{A}h(t) + \mathbf{B}x(t),  \\
  y(t)  = \mathbf{C}h(t) + \mathbf{D}x(t).
\end{split},
\end{equation}
with evolution parameter matrix $\mathbf{A} \in \mathbb{C}^{N \times N}$,
projection parameters $\mathbf{B}, \mathbf{C} \in \mathbb{C}^{N}$, defined for a state dimension $N$, alongside a skip connection parameter $\mathbf{D} \in \mathbb{C}^{1}$.

Analogous to previous works \cite{mamba,vim}, our encoder implements a discretized version of this continuous system that aligns it to the digital nature of the input data \cite{gu2021combining}. 
Given a discrete input $x_k \in \mathbb{R}^{L \times D}$, the discretization of Eq.~\ref{eq:ssm} under the zeroth-order hold (ZOH) method using a step size $\Delta$ yields:
\begin{align}\label{eqn:discrete}
\begin{split}
\mathbf{\bar{A}} = e^{\Delta \mathbf{A}}, \quad
\mathbf{\bar{B}} &\approx (\Delta \mathbf{A})(\Delta \mathbf{A})^{-1}\Delta \mathbf{B} = \Delta \mathbf{B}, \quad
\mathbf{\bar{C}} = \mathbf{C}\\
h_k &= \mathbf{\bar{A}}h_{k-1} + \mathbf{\bar{B}}x_k, \\ 
y_k &= \mathbf{\bar{C}}h_k + \mathbf{\bar{D}}x_k, \\
\end{split}
\end{align}
where we consider the first-order Taylor approximation for $\mathbf{\bar{B}}$ defined in \cite{mamba}.
In S6 \cite{mamba} and our case, the matrices $\mathbf{B} \in \mathbb{R}^{B \times L \times N}$, $\mathbf{C} \in \mathbb{R}^{B \times L \times N}$, and $\Delta \in \mathbb{R}^{B \times L \times D}$ are dynamically derived from the input data $x \in \mathbb{R}^{B \times L \times D}$.

While classical SSM formulations handle 1D sequences, we follow the approach of Vim \cite{vim} (and VMamba \cite{liu2024vmamba}) to process 2D feature maps by scanning row-wise (or column-wise). Specifically, at each encoder block, we reshape the $H \times W \times C$ feature map into a sequence of length $H \times W$ and channel dimension $C$, apply the SSM transformation, and then reshape back. Repeated blocks with positional embeddings ensure the model can capture 2D global context despite operating on 1D scans.

P-SSE is constructed by repeating the block structure in a ViT fashion, similar to \cite{Zubic_2024_CVPR}. These features constitute the output of our P-SSE encoders, which we use to extract correlation and context features as described in the Sec. \ref{sec:model_arch}. To ensure the stability of the training, we first learn a perturbation matrix for the state matrix on the ImageNet dataset \cite{deng_cvpr_2009}, and then use it, along with a perturb-then-diagonalize (PTD) technique, during training in the same fashion as \cite{yu2024robustifying}. However, they tested the model only on synthetic tasks.
We use the $L1$ loss between the predicted and ground truth bidirectional optical flows to supervise the training of our model \cite{zubic_aiai_2021, Shi_2023_ICCV}.
\section{Explanation of Perturb-then-Diagonalize (PTD) Technique}
\label{suppl:ptd_technique}
\subsection{Preliminaries}
The Perturb-then-Diagonalize (PTD) technique \cite{yu2024robustifying} aims to enhance the performance and stability of SSMs by making the state dynamics matrix $\mathbf{A}$ more robust to data noise, surpassing even the HiPPO theory-based matrix initialization \cite{gu2020hippo}. This method is particularly effective for high-dimensional data and complex dynamics, as encountered in our case. The PTD method comprises two primary steps applied to the state dynamics matrix $\mathbf{A}$ of an SSM:. In the first step, the system matrix $\mathbf{A}$ is perturbed by a small matrix $\mathbf{E}$, yielding a new matrix $\mathbf{A}^*$. The perturbation matrix $\mathbf{E}$ is carefully chosen to enhance system properties like stability and robustness without significantly altering its original behavior. Following \cite{yu2024robustifying}, we set the magnitude of $\mathbf{E}$ to approximately 10\% of $\mathbf{A}$'s magnitude. This process is mathematically represented as:
\begin{equation}
\mathbf{A}^* = \mathbf{A} + \mathbf{E},
\end{equation}
where $\mathbf{E}$ is typically small in magnitude compared to $\mathbf{A}$. We constrain the size of the perturbation matrix to 10\% of the size of the HiPPO \cite{gu2020hippo} matrix $\mathbf{A}$. Following perturbation, the perturbed matrix $\mathbf{A}^*$ is diagonalized.

\subsection{Application of the Technique to Our Problem}
Incorporating pre-training in deep learning frameworks improved the model performance \cite{Zubic_2023_ICCV}, particularly in tasks requiring high-level feature recognition. Our work leverages the PTD \cite{yu2024robustifying} technique to pre-train a model on ImageNet \cite{deng_cvpr_2009}, a large-scale image classification dataset, focusing specifically on the learned $\mathbf{A}$ matrix.

We initiate by pre-training our model using the PTD \cite{yu2024robustifying} technique for SSMs on ImageNet \cite{deng_cvpr_2009}. PTD introduces controlled perturbations in the training dynamics, enhancing the model's ability to escape local minima and improving generalization. We modify the loss function $L_{\text{PTD}}(A)$ for the $\mathbf{A}$ matrix as follows:
\begin{equation}
L_{\text{PTD}}(A) = L(A) + \lambda \cdot |\mathbf{E}|,
\end{equation}
where $|\mathbf{E}|$ represents the magnitude of perturbations in the model's parameters $\theta$, and $\lambda$ controls these perturbations' strength. As previously stated, the perturbation magnitude is approximately 10\% of the $\mathbf{A}$ matrix magnitude. Perturbations $\mathbf{E}$ are sampled from a Gaussian distribution \cite{yu2024robustifying}, encouraging exploration of a wider parameter space during training.

Post-ImageNet \cite{deng_cvpr_2009} pre-training, we utilize the state dynamics matrix $\mathbf{A}^*$ learned within this PTD framework to initialize our state dynamics matrix. This initialization is then used to train the model on the DSEC dataset \cite{Gehrig21ral} and perform a zero-shot evaluation on the MVSEC dataset \cite{zhu_ral_2018}.
\section{Experiments}\label{chap:experiments}

\textbf{Datasets and Evaluation Metric.} We train our models on the DSEC-Flow dataset \cite{Gehrig21ral} and evaluate them on both DSEC and MVSEC \cite{zhu_ral_2018} (zero-shot) datasets. DSEC-Flow provides 7800 training instances, each with bidirectional optical flow ground-truth. In Table~\ref{tab:dsec_flow_performance}, we report results obtained from the official DSEC-Flow server. Notably, the raw training event data is initially distorted and not rectified, necessitating pre-processing to rectify events before model training, ensuring compatibility with the rectified event representations used in benchmark tests.

MVSEC \cite{zhu_ral_2018} comprises 5 driving sequences and 4 indoor drone-collected sequences, with ground truth derived from lidar odometry and mapping. Following \cite{wan2022learning}, we focus primarily on the \emph{indoor\_flying1}, \emph{indoor\_flying2}, and \emph{indoor\_flying3} sequences to demonstrate our model's generalization capabilities on out-of-distribution data. In accordance with benchmark protocols \cite{Gehrig21ral}, we primarily use End-Point Error (EPE) to evaluate optical flow accuracy, calculating the mean L2 distance between predicted and actual flow vectors. The DSEC benchmark additionally employs Angular Error (AE) and N-pixel Error (NPE), the latter quantifying the proportion of pixels where optical flow error exceeds a given threshold $N \in \{1,2,3\}$.

\textbf{Implementation Details.} Our models are implemented using PyTorch. We primarily utilize the ERGO-12 representation \cite{Zubic_2023_ICCV} with 12 bins. Feature and context embeddings are generated using our proposed P-SSE. For motion feature estimation and bidirectional flow learning, we employ the GMA-SK update block \cite{Sun_NeurIPS}. Both E-TROF and E-MOP architectures undergo 250k training iterations with batch size 4. We use the AdamW optimizer with a $4\times10^{-4}$ learning rate and a OneCycle LR Scheduler. We apply data augmentation techniques (horizontal/vertical flips, random cropping) for robustness. Unless otherwise stated, training and inference are conducted on either A100 or V100 GPUs.

\subsection{DSEC-Flow Results}\label{sec:dsec}
In Table~\ref{tab:dsec_flow_performance}, we compare our approach to a range of existing event-based optical flow methods, including TMA~\cite{Liu_2023_ICCV}, BFlow~\cite{Gehrig2024pami}, IDNet~\cite{idnet}, and ECDDP~\cite{Yang2024EventCD}. Although IDNet and ECDDP were published concurrently or shortly after our initial development, we highlight their performance for completeness. As shown, our P-SSE achieves an EPE of 0.68, which is lower than IDNet’s 0.723 and ECDDP’s 0.697, marking an improvement of approximately 6\% compared to IDNet and about 2.4\% compared to ECDDP. Additionally, our method outperforms TMA (0.743 EPE) and BFlow (0.750 EPE). Our approach is also the only one in Table~\ref{tab:dsec_flow_performance} that integrates a multi-event design within a perturbed state space model. 

\begin{table}[t]
\centering
\setlength{\tabcolsep}{3pt}
\caption{Performance comparison on the DSEC-Flow Public Benchmark \cite{Gehrig21ral}. 1PE, 2PE, 3PE are the N-pixel errors. The top result in each column is in bold.}
\label{tab:dsec_flow_performance}
\begin{tabular}{lccccc}
\toprule
Method & 1PE & 2PE & 3PE & EPE & AE \\
\midrule
E-RAFT \cite{Gehrig3dv2021} & 12.742 & 4.740 & 2.684 & 0.788 & 2.851 \\
ADMFlow \cite{Luo_2023_ICCV} & 12.522 & 4.673 & 2.647 & 0.779 & 2.838 \\
E-FlowFormer \cite{blinkflow_iros2023} & 11.225 & 4.102 & 2.446 & 0.759 & 2.676 \\
BFlow \cite{Gehrig2024pami} & 11.901 & 4.411 & 2.440 & 0.750 & 2.680 \\
TMA \cite{Liu_2023_ICCV} & 10.863 & 3.972 & 2.301 & 0.743 & 2.684 \\
IDNet \cite{idnet} & 10.111 & 3.523 & 2.018 & 0.723 & 2.724 \\
ECDDP \cite{Yang2024EventCD} & \textbf{8.887} & \textbf{3.199} & 1.958 & 0.697 & 2.575 \\
\textbf{Ours (P-SSE)} & 9.144 & 3.232 & \textbf{1.816} & \textbf{0.680} & \textbf{2.560} \\
\bottomrule
\end{tabular}
\end{table}

\begin{figure*}[t]
    \centering
    \includegraphics[width=0.8\textwidth]{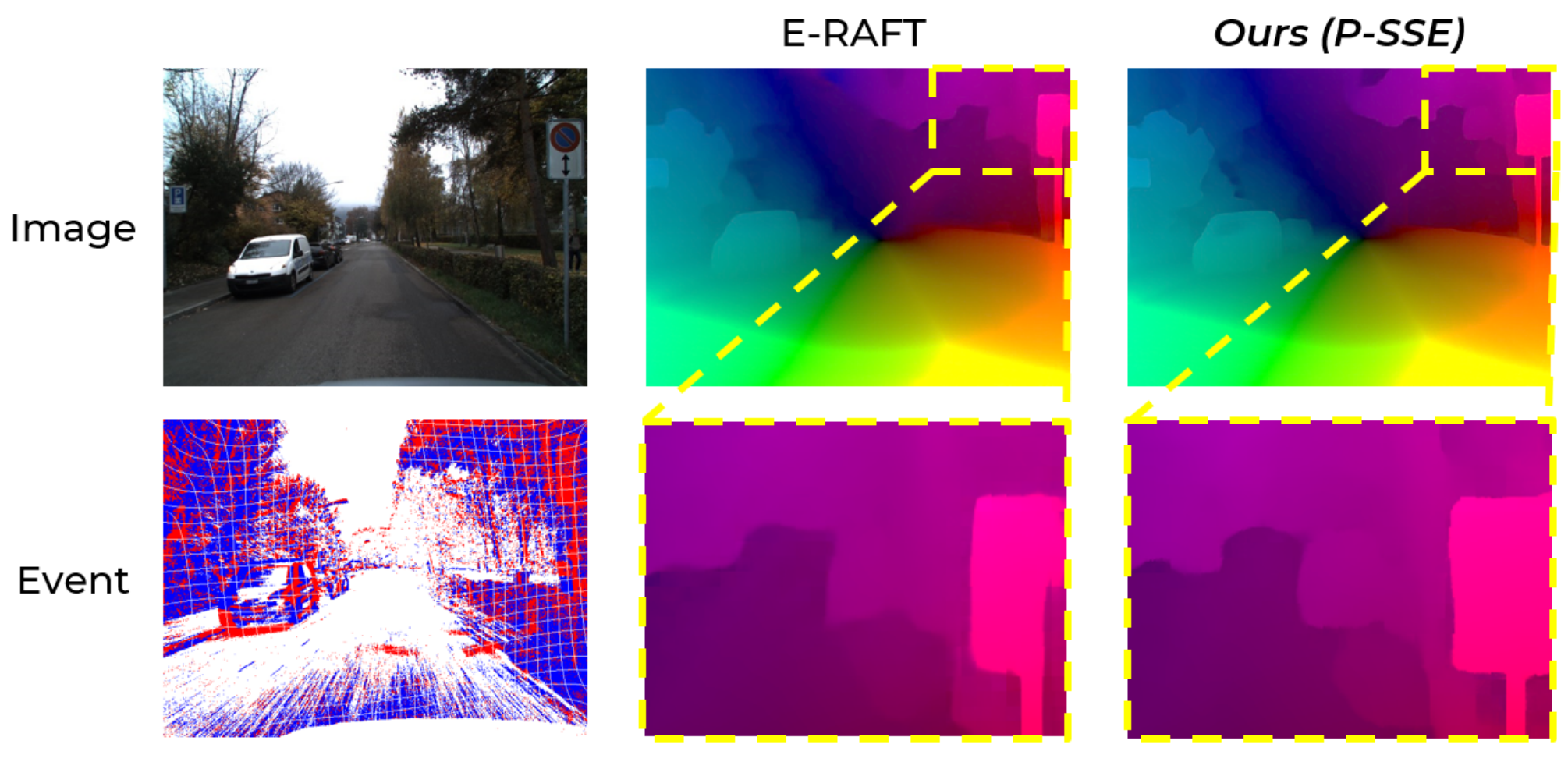}
    \caption{Illustration of P-SSE's efficacy in handling out-of-boundary regions, in comparison with E-RAFT~\cite{Gehrig3dv2021}.}
    \label{img:oob}
\end{figure*}

\begin{figure}[t]
    \centering
    \includegraphics[width=\columnwidth]{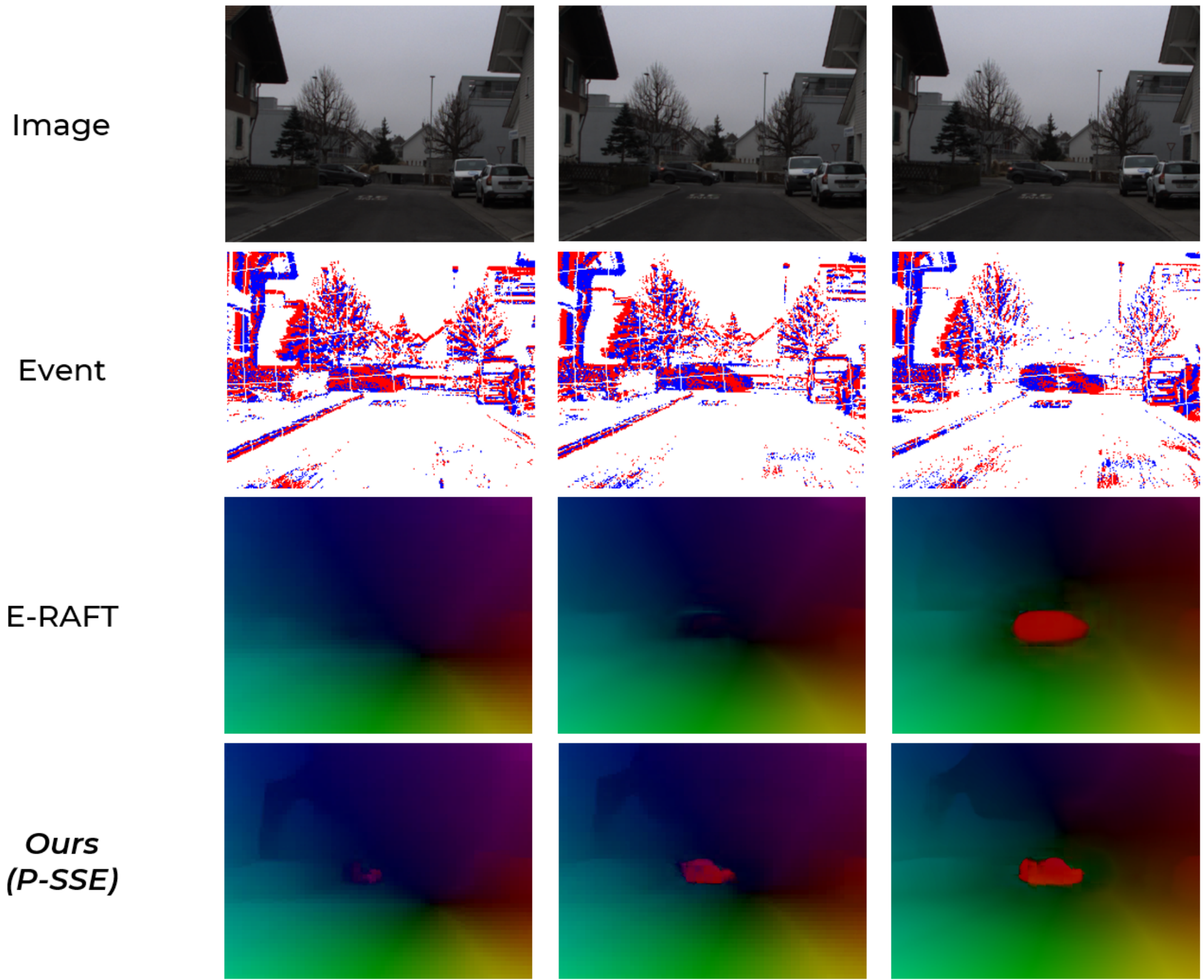}
    \caption{Demonstration of P-SSE's capability in managing partially occluded scenes, compared with E-RAFT~\cite{Gehrig3dv2021}. The sequence progresses from left to right, showcasing frames from a DSEC-Flow test sequence.}
    \label{img:occlusion}
\end{figure}

Fig.~\ref{img:oob} demonstrates how our model excels in predicting flow for a street sign moving partially out of frame, where E-RAFT~\cite{Gehrig3dv2021} struggles. Similarly, Fig.~\ref{img:occlusion} shows that P-SSE reliably tracks a car even when it is partially occluded. We attribute these improvements to the stability of our state matrix (enabled by perturbation), which helps effectively leverage temporal information across multiple events. Additionally, Section~\ref{sec:analysis} discusses the resource efficiency of our model in more detail, demonstrating that our multi-event design does not significantly inflate training/inference costs compared to lighter models such as IDNet.

\subsection{MVSEC Results}\label{sec:multiflow}
Table~\ref{tab:mvsec_results} presents performance on the MVSEC~\cite{zhu_ral_2018} \emph{indoor\_flying} sequences under the protocol of \cite{wan2022learning} (with $dt = 1$). To test generalization, we train on DSEC and evaluate on MVSEC without fine-tuning, following \cite{wan2022learning}. This setting is challenging due to substantial domain differences: DSEC covers driving scenarios with a DAVIS346 sensor, while MVSEC’s indoor sequences use a DAVIS240. Despite these disparities, our network outperforms existing methods and achieves a remarkable 11.86\% improvement over the previous best results. 

\begin{table}[t]
\centering
\small  
\setlength{\tabcolsep}{2.5pt} 
\caption{EPE results on MVSEC. “in\_fly” = “indoor\_flying”. We report results for both \emph{Base} and \emph{Small} variants of P-SSE using either ERGO-12 or Time Surface representations.}
\label{tab:mvsec_results}
\begin{tabular}{lllcccc}
\hline
\multicolumn{3}{l}{Methods} & \textit{in\_fly1} & \textit{in\_fly2} & \textit{in\_fly3} & $\overline{\textit{EPE}}$ \\ \hline
\multicolumn{3}{l}{EV-FlowNet \cite{Zhu-RSS-18}}  
    & 1.03 & 1.72 & 1.53 & 1.43 \\
\multicolumn{3}{l}{Zhu et al. \cite{Zhu19cvpr}}   
    & 0.58 & 1.02 & 0.87 & 0.82 \\
\multicolumn{3}{l}{EST \cite{gehrig2019end}}      
    & 0.97 & 1.38 & 1.43 & 1.26 \\
\multicolumn{3}{l}{Matrix-LSTM \cite{cannici2020differentiable}} 
    & 0.82 & 1.19 & 1.08 & 1.03 \\
\multicolumn{3}{l}{Spike-FlowNet \cite{lee2020spike}}             
    & 0.84 & 1.28 & 1.11 & 1.08 \\
\multicolumn{3}{l}{Stoffregen et al. \cite{stoffregen2020reducing}} 
    & 0.56 & 0.66 & 0.59 & 0.60 \\
\multicolumn{3}{l}{Paredes et al. \cite{paredes2021back}}         
    & 0.79 & 1.40 & 1.18 & 1.12 \\
\multicolumn{3}{l}{LIF-EV-FlowNet \cite{hagenaars2021self}}       
    & 0.71 & 1.44 & 1.16 & 1.10 \\
\multicolumn{3}{l}{Deng et al. \cite{deng2021learning}}           
    & 0.89 & 0.66 & 1.13 & 0.89 \\
\multicolumn{3}{l}{Li et al. \cite{li2021lightweight}}            
    & 0.59 & 0.64 & -    & 0.62 \\
\multicolumn{3}{l}{STE-FlowNet \cite{ding2022spatio}}             
    & 0.57 & 0.79 & 0.72 & 0.69 \\
\multicolumn{3}{l}{DCEIFlow \cite{wan2022learning}}               
    & 0.56 & 0.64 & 0.57 & 0.59 \\

\multicolumn{3}{l}{\textbf{Ours (P-SSE)}:}\\
& \multirow{2}{*}{ERGO-12}  & Base   
    & \textbf{0.47} & 0.59 & \textbf{0.50} & \textbf{0.52} \\
&                           & Small  
    & 0.52 & 0.63 & 0.56 & 0.57 \\
& \multirow{2}{*}{Time Surf} & Base  
    & \textbf{0.47} & 0.58 & \textbf{0.50} & \textbf{0.52} \\
&                           & Small  
    & 0.48 & \textbf{0.57} & 0.51 & \textbf{0.52} \\
\hline
\end{tabular}
\end{table}

\subsection{Case Study and Analysis}\label{sec:analysis}

\noindent\textbf{Train and Inference Efficiency.} 
Table~\ref{tab:count_flops} compares our optical flow prediction network against other state-of-the-art architectures, including both image-based (RAFT~\cite{Teed2020RAFTRA}) and event-based (TMA~\cite{Liu_2023_ICCV}, BFlow~\cite{Gehrig2024pami}) approaches. All measurements are taken on an NVIDIA Quadro RTX 8000 with batch size 1, averaged over 100 runs. Our full model (\emph{P-SSE\textsubscript{base}}) achieves an inference time of 111.2\,ms, while a \emph{small} (4.12\,M parameters) and \emph{nano} (2.99\,M) variant run even faster at 100.4\,ms and 95.1\,ms respectively, maintaining competitive accuracy.

\begin{table}[t]
\centering
\caption{Efficiency Comparison on Quadro RTX 8000 GPU.}
\label{tab:count_flops}
\setlength{\tabcolsep}{0.4em}
\begin{tabular}{lcccc}
\hline
\multirow{2}{*}{Methods} & Inference & \# of Params & VRAM \\
                         & Time [ms] & [M]          & [GB]  \\ \hline
RAFT \cite{Teed2020RAFTRA}       & 226.4 & 5.26 & \textbf{4.35} \\
TMA \cite{Liu_2023_ICCV}         & 201.4 & 6.88 & 10.03 \\
BFlow \cite{Gehrig2024pami}      & 133.2 & 5.64 & 7.93  \\
\midrule
\textbf{(Ours) P-SSE\textsubscript{base}} & 111.2 & 6.01 & 12.0  \\
\textbf{(Ours) P-SSE\textsubscript{small}} & 100.4 & 4.12 & 10.0  \\
\textbf{(Ours) P-SSE\textsubscript{nano}}  & \textbf{95.1} & \textbf{2.99} & 8.3  \\
\hline
\end{tabular}
\end{table}

\vspace{0.5em}
\noindent\textbf{Perturb-Then-Diagonalize Strategy.}
In Table~\ref{tab:method_epe_comparison}, we evaluate the impact of the PTD~\cite{yu2024robustifying} technique versus the standard HiPPO~\cite{gu2020hippo} initialization. We observe that PTD yields both higher training stability (fewer spikes in validation loss) and a consistent reduction in final EPE (3.09\% improvement for \emph{small}, 2.41\% for \emph{base}). These results suggest that small perturbations to the state matrix improve the convergence properties of our SSM-based encoder and are especially beneficial for event-based optical flow, which typically exhibits sparse and highly asynchronous inputs.

\begin{table}[t]
\centering
\caption{Comparison of methods using PTD (vs.\ HiPPO) for EPE on DSEC-Flow.}
\label{tab:method_epe_comparison}
\setlength{\tabcolsep}{5pt}
\begin{tabular}{clc}
\toprule
Methods & Initialization & EPE \\
\midrule
\multirow{2}{*}{P-SSE\textsubscript{Small}} & PTD \cite{yu2024robustifying}  & 0.691 \\
                                           & HiPPO \cite{gu2020hippo}       & 0.713 \\
\multirow{2}{*}{P-SSE\textsubscript{Base}} & PTD \cite{yu2024robustifying}  & 0.680 \\
                                           & HiPPO \cite{gu2020hippo}       & 0.705 \\
\bottomrule
\end{tabular}
\end{table}
\section{Conclusion}
\label{sec:conclusion}
This work introduces a novel approach to optical flow estimation using event-based cameras, significantly improving the temporal and spatial reasoning crucial for accurate and efficient flow computation. P-SSE improves spatial feature encoding by combining the global receptive field of Transformers \cite{nips2017_vaswani} with the efficiency of SSMs. We implement the PTD technique \cite{yu2024robustifying} in our P-SSE to ensure model stability, effectively addressing the challenges of training deep networks on complex event-based data. Inspired by recent advances in frame-based optical flow \cite{Shi_2023_ICCV}, we adopt a multi-frame approach that utilizes bi-directional flows and propagates motion information across multiple time instants. This extension of the temporal receptive field significantly enriches the model's motion understanding.

Empirical evaluations on DSEC-Flow \cite{Gehrig21ral} and MVSEC \cite{zhu_ral_2018} datasets demonstrate the superiority of our approach. We not only establish new benchmarks in optical flow estimation accuracy but also achieve substantial improvements in computational efficiency. The notable reductions in training time and resource requirements, coupled with faster inference, underscore the practicality and applicability of our method in real-world scenarios, particularly in domains demanding rapid and reliable motion estimation.

\section{Acknowledgment}
\label{sec:acknowledgment}
This work was supported by the European Research Council (ERC) under grant agreement No. 864042 (AGILEFLIGHT).

{
    \small
    \bibliographystyle{ieeenat_fullname}
    \bibliography{main}
}

\end{document}